\documentclass[sigconf]{acmart}

\AtBeginDocument{%
  \providecommand\BibTeX{{%
    \normalfont B\kern-0.5em{\scshape i\kern-0.25em b}\kern-0.8em\TeX}}}

\acmPrice{15.00}
\acmISBN{978-1-4503-XXXX-X/18/06}

\usepackage{amsmath,amsfonts,bm}

\def\eqref#1{equation~\ref{#1}}

\def\1{\bm{1}}

\def\mW{{\bm{W}}}

\DeclareMathAlphabet{\mathsfit}{\encodingdefault}{\sfdefault}{m}{sl}
\SetMathAlphabet{\mathsfit}{bold}{\encodingdefault}{\sfdefault}{bx}{n}

\newcommand{\E}{\mathbb{E}}

\newcommand{\softmax}{\mathrm{softmax}}

\usepackage{hyperref}
\usepackage{url}

\usepackage{caption}
\usepackage{subcaption}
\usepackage{graphicx}

\usepackage{multirow}
\usepackage{booktabs}
\usepackage{subcaption}
\usepackage{nicematrix,tikz,makecell}
\usepackage{wrapfig}
\usepackage{amsthm}
\usepackage{amsmath}
\usepackage{amsthm}
\usepackage{amsfonts}
\usepackage{bbm}
\usepackage{bm}
\usepackage{mathtools}

\usepackage{bbding} 

\usepackage{xspace}
\newcommand{\our}{UniLP\xspace}

\usepackage{color, colortbl}
\definecolor{Gray}{gray}{0.9}

\usepackage{xcolor}        
\newcommand{\firstt}[2]{\textcolor{red}{$\mathbf{{#1} {\scriptstyle \pm {#2}}}$}}%
\newcommand{\secondt}[2]{\textcolor{blue}{$\mathbf{{#1} {\scriptstyle \pm {#2}}}$}}%
\newcommand{\thirdt}[2]{\textcolor{violet}{$\mathbf{{#1} {\scriptstyle \pm {#2}}}$}}%

\newcommand{\firstts}[1]{\textcolor{red}{$\mathbf{{#1}}$}}%
\newcommand{\secondts}[1]{\textcolor{blue}{$\mathbf{{#1}}$}}%
\newcommand{\thirdts}[1]{\textcolor{violet}{$\mathbf{{#1}}$}}%

\definecolor{MyRed}{HTML}{EA6B66}
\definecolor{MyGreen}{HTML}{97D077}

\theoremstyle{plain}
\newtheorem{theorem}{Theorem}[section]

\theoremstyle{definition}
\newtheorem{definition}[theorem]{Definition}

\theoremstyle{remark}

\newcommand{\kevin}[1]{\textcolor{red}{KD:~#1}}
\newcommand{\mao}[1]{\textcolor{blue}{mao:~#1}}
\newcommand{\maomask}[1]{}

\renewcommand{\kevin}[1]{}
\renewcommand{\mao}[1]{}

\begin{document}

\title{Universal Link Predictor By In-Context Learning on Graphs}

\author{Kaiwen Dong}
\affiliation{\institution{University of Notre Dame}\country{USA}}
\email{kdong2@nd.edu}

\author{Haitao Mao}
\affiliation{\institution{Michigan State University}\country{USA}}
\email{haitaoma@msu.edu}

\author{Zhichun Guo}
\affiliation{\institution{University of Notre Dame}\country{USA}}
\email{zguo5@nd.edu}

\author{Nitesh V. Chawla}
\affiliation{\institution{University of Notre Dame}\country{USA}}
\email{nchawla@nd.edu}

\begin{abstract}
Link prediction is a crucial task in graph machine learning, where the goal is to infer missing or future links within a graph. Traditional approaches leverage heuristic methods based on widely observed connectivity patterns, offering broad applicability and generalizability without the need for model training. Despite their utility, these methods are limited by their reliance on human-derived heuristics and lack the adaptability of data-driven approaches. Conversely, parametric link predictors excel in automatically learning the connectivity patterns from data and achieving state-of-the-art but fail short to directly transfer across different graphs. Instead, it requires the cost of extensive training and hyperparameter optimization to adapt to the target graph. In this work, we introduce the Universal Link Predictor (\our), a novel model that combines the generalizability of heuristic approaches with the pattern learning capabilities of parametric models. \our is designed to autonomously identify connectivity patterns across diverse graphs, ready for immediate application to any unseen graph dataset without targeted training. We address the challenge of conflicting connectivity patterns—arising from the unique distributions of different graphs—through the implementation of In-context Learning (ICL). This approach allows \our to dynamically adjust to various target graphs based on contextual demonstrations, thereby avoiding negative transfer. Through rigorous experimentation, we demonstrate \our's effectiveness in adapting to new, unseen graphs at test time, showcasing its ability to perform comparably or even outperform parametric models that have been finetuned for specific datasets. Our findings highlight \our's potential to set a new standard in link prediction, combining the strengths of heuristic and parametric methods in a single, versatile framework.

\end{abstract}




\maketitle

\section{Introduction}

Graph-structured data is ubiquitous across diverse domains, including social networks~\cite{liben-nowell_link_2003}, protein-protein interactions~\cite{szklarczyk_string_2019}, movie recommendations~\cite{koren_matrix_2009}, and citation networks~\cite{yang_revisiting_2016}. It encapsulates the complex relationships among entities, serving as a powerful data structure for analytical exploration. At the heart of graph analysis lies the task of link prediction (LP)~\cite{yang_evaluating_2015,dong_fakeedge_2022, guo_linkless_2022}, a crucial problem aimed at forecasting missing or future connections within these networks. Over the years, the quest to enhance LP accuracy has advanced the development of numerous methodologies~\cite{kumar_link_2020}, broadly categorized into two main classes of approaches.

The first line of works is non-parametric heuristics link predictors, including Common Neighbor (CN)~\cite{liben-nowell_link_2003}, Preferential Attachment (PA)~\cite{barabasi_emergence_1999}, Resource Allocation (RA)~\cite{zhou_predicting_2009} and Katz index~\cite{katz_new_1953}. By discovering and abstracting the universal structural properties underlying different graphs~\cite{barabasi_emergence_1999,watts_collective_1998,holland_stochastic_1983}, heuristics methods are developed based on observing the connectivity patterns existing in real-world graph datasets. For example, CN assumes the tendency of triadic closure~\cite{easley_networks_2010}, such that a friend's friend is likely to be friends in a social network. These heuristics link predictors can be readily applied to any graph dataset with great generalizability. 
However, this approach relies on predefined heuristics, crafted from human expertise into the graph connectivity. Despite the initial success via capturing one specific connectivity pattern, they fail to capture all the effective structural features in the link prediction, leading to suboptimal performance when applied indiscriminately.

The other line of works is parametric link predictors, which automatically learn the connectivity patterns by fitting the LP models to the target graphs. These parametric methods, especially those Graph Neural Networks~\cite{kipf_semi-supervised_2017,hamilton_inductive_2018} for Link Prediction~(GNN4LP), have dominated the leaderboard of the link prediction tasks~\cite{hu_open_2021}. Typically, these GNN4LP are provably the most expressive models such that the link representation is permutation-invariant~\cite{zhang_labeling_2021}. They can capture more effective structural features compared to the simpler heuristics counterparts. However, their dependency on extensive training for each new graph dataset and the necessity for hyperparameter optimization~\cite{chamberlain_graph_2022,wang_neural_2023,dong_pure_2023} present notable challenges for their application across diverse graph environments.

Given that (1) the heuristics methods can be readily applied to any graphs without training based on common connectivity patterns and (2) the parametric model can automatically capture the connectivity patterns by fitting on the graph, a natural question arises:
\begin{center}
    \textbf{\textit{Can a singular LP model automatically learn and apply the connectivity pattern across new, unseen graphs without the need for direct training?}}
\end{center}
An affirmative response would not only pioneer a new frontier in graph machine learning but also align with the transformative potential observed in foundation models across text and image processing fields~\cite{brown_language_2020,kirillov_segment_2023}. These models' exceptional generalizability, driven by their capability of transfer learning~\cite{yosinski_how_2014}, offers a blueprint for the development of a universal LP model capable of broad applicability without explicit fitting.


\paragraph{Present work.} In this study, we introduce \maomask{I do not quite understand what is the universal link predictor}the Universal Link Predictor (\our), a novel model designed for immediate application across diverse graph environments\footnote{In this study, we focus on non-attributed graphs. This choice is informed by previous findings indicating that node attributes have minimal impact on the effectiveness of LP tasks~\cite{dong_pure_2023}. \maomask{I think another thing is that the feature can be hardly transferable. It may be a good choice to add one specific feature channel for each downstream task. Nonetheless, such specific rather than transferability is not the focus of our current paper}} without the prerequisite of model fitting. Our investigation starts by assessing whether existing LP models possess the capability to transfer connectivity pattern knowledge from one graph to another. \maomask{I think it could be better to say how we specilize the most-expressive structural representation for the given downstream task. That means, we know all the patterns from GNN, what pattern to utilize is specific on each dataset and selected by the ICL.} Through empirical and theoretical analyses spanning both heuristic and parametric link predictors, we uncover a significant challenge: negative transfer~\cite{wang_afec_2021} can happen when directly transferring the connectivity patterns across distinct graph datasets, including both real-world and synthetic examples. This complexity arises from the inherent diversity and flexibility of graph data, leading to unique connectivity patterns for each graph.

To equip \our with the capability to adapt to diverse graphs without the need for training, we are inspired by the concept of In-context Learning (ICL) as utilized by large language models (LLMs)~\cite{brown_language_2020}. ICL enables models to adapt to new datasets or tasks through the relevant demonstration examples~\cite{wang_label_2023}. Analogously, for adapting our LP model to a particular graph, we select a collection of in-context links to act as such demonstration examples. These in-context links not only provide a context for link prediction but also aid in capturing the unique connectivity pattern inherent to the graph in question. To achieve link representations that are conditioned on the graph’s specific connectivity pattern, we employ an attention mechanism~\cite{vaswani_attention_2017,brody_how_2022}. This mechanism facilitates dynamic adjustment of link representations in response to the graph context, enabling the model to accurately reflect the unique connectivity patterns of each graph.

We have curated a diverse collection of graph datasets spanning multiple domains, providing a rich variety of connectivity patterns for benchmarking. Through extensive experiments on these datasets, we demonstrate the seamless applicability of \our to novel and unseen graph datasets without requiring dataset-specific fitting. Notably, \our, empowered with ICL, exhibits the capability to meet or even exceed the performance levels of LP models that have been pretrained and finetuned for specific target graphs. This achievement underscores \our's broad applicability and robust adaptability, establishing a groundbreaking approach to link prediction tasks.

In summary, our contributions to the field of link prediction are:
\begin{itemize}
\item We pioneer in highlighting the challenges of applying a singular LP model across various graph datasets due to conflicting connectivity patterns, a finding supported by both empirical evidence and theoretical analysis.
\item Addressing these challenges, we introduce \our, a novel LP approach leveraging ICL for dynamic adaptation to new graphs in real-time, thereby eliminating the need for traditional training processes.
\item The diverse collection of graph datasets we've collected facilitates extensive validation of \our's adaptability. Our experiments confirm that \our is not only capable of adjusting to any new graph dataset during inference but also achieves competitive performance, marking a significant advancement in link prediction methodologies.
\end{itemize}




%


\section{Can one model fit all?} 
\mao{I think you may consider to change the title into a more concrete one: Can one GNN4LP perform well across graphs from different domains.}

Machine learning models perform a task by learning from data. The quest for generalizability in machine learning models has led to significant advancement in domains such as natural language processing (NLP)~\cite{vaswani_attention_2017,brown_language_2020} and computer vision (CV)~\cite{betker_improving_nodate,kirillov_segment_2023}. Foundation models in these fields have demonstrated remarkable generalizability across unseen datasets~\cite{donahue_decaf_2014}, primarily due to their training on extensive data, which enables them to learn transferable knowledge. 

In the context of LP tasks, \maomask{the factor (or the principle) can be transfer, refer to my paper, not heuritisc, heuritisc is just observation}the heuristics link predictors can be seen as a type of \emph{transferable knowledge}. These predictors, crafted by manually analyzing common connectivity patterns in real-world graphs, offer insights into the underlying structure of networks. However, the validity of applying these heuristics universally is questioned, especially considering the wide spectrum of graph data. For instance, social networks like Facebook often exhibit a community-oriented structure~\cite{newman_modularity_2006}. Conversely, networks adhering to a scale-free power-law distribution~\cite{barabasi_emergence_1999}, such as the World Wide Web, tend to favor a Preferential Attachment connectivity pattern. \mao{I think the example may be change to my introduction one which shows an obvious conflict} Through both empirical and theoretical examination, we aim to explore the challenges posed by the direct application of connectivity patterns \mao{I think the connectivity patterns is not well defined } from one graph to another. Our findings will reveal that such an approach may lead to negative transfer~\cite{wang_afec_2021}, emphasizing the critical need for adaptable strategies in the face of graph diversity.
\mao{add a short organization: In section 2.1, we xxxxx}


\subsection{Empirical evaluation on transferability}
Our exploration begins with an empirical investigation aimed at understanding the transferability of learned connectivity patterns across diverse graph domains. We curate a collection of real-world graphs from varied fields in Table~\ref{tab:stats}, ensuring comprehensive illustrations of different graph types. 

To assess the potential of the important connectivity pattern, learned from one graph, to influence the LP performance on another, we incorporate extra graphs into the training phase of the target graph. In other words, this experiment deviates from the standard supervised learning approach by introducing additional training signals from other graphs. 
If the connectivity patterns from these extra graphs align with or augment the structure of the target graph, the LP model's performance should either remain stable or improve. To make the experiment tractable, we only introduce \textbf{one additional} graph into the training graph and then make a link prediction on the target graph. This additional graph is kept disconnected from the target graph to ensure that the test set remains the same as standard LP tasks. \mao{I do not quite understand the above sentence}. In these experiments, we employ SEAL~\cite{zhang_link_2018} as the backbone model for the experiment and adopt Hits@50 as the performance metrics~\cite{hu_open_2021}. \mao{We may add an additional words that the test remains the same?}

\begin{figure}[h]
    \centering
    \resizebox{1.0\linewidth}{!}{%
    \includegraphics[width=1.0\linewidth]{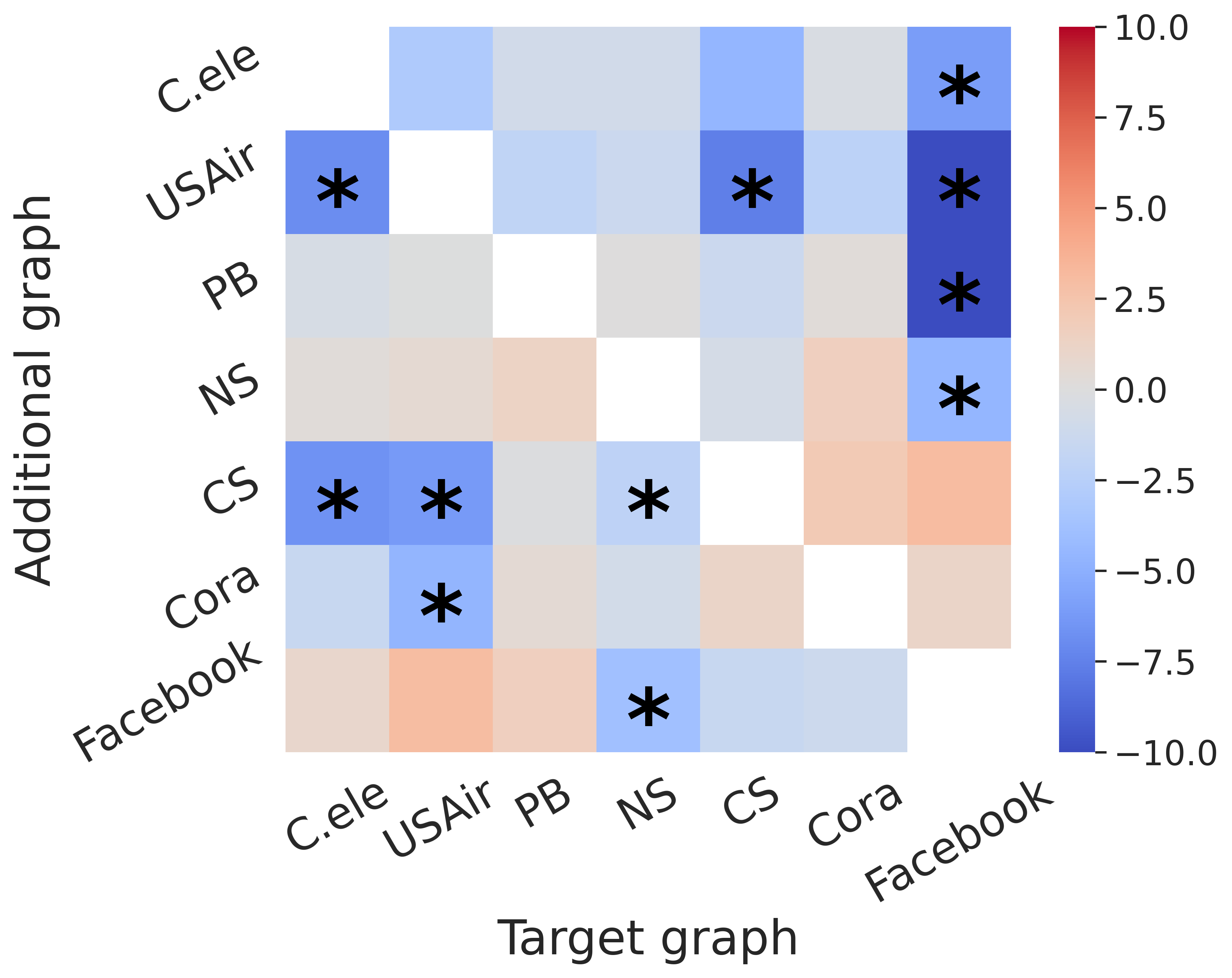}
    }
    \caption{Performance change of SEAL~\cite{zhang_link_2018} after training with one additional graph. \AsteriskBold denotes statistically significant change.}
    \label{fig:transfer}
\end{figure}

\sloppy
Results are presented in Figure~\ref{fig:transfer}. It shows how the LP model's performance is affected by the introduction of additional graph data during training. In the heatmap, warm colors represent an improvement in LP performance, while cooler colors denote a performance decrease. The predominance of cooler colors in the heatmap reveals that integrating an extra graph into training generally results in performance degradation. This observation underscores the potential discordance in the underlying characteristics of different graphs, leading to conflicts between the learned connectivity patterns \mao{the learned pattern is the effective pattern on link prediction}. This phenomenon highlights the inherent challenge in deploying a singular LP model across various graphs, thus questioning the feasibility of a ``one model fits all'' approach in the context of LP tasks.
\mao{Add one sentence transition to the following paragraph: to delve deep into the underlying reason on the negative transfer, xxxx}

\subsection{Conflicting patterns across graphs}
In this section, we delve into the theoretical aspects of how the unique characteristics of different graphs can hinder the transferability of connectivity patterns. We begin with a formal definition and preliminary discussion of LP.

\paragraph{Preliminary.} Consider an undirected graph $G=(V,E^o)$. $V$ is the set of nodes with size $n$, which can be indexed as $\{i\}_{i=1}^{n}$. $E^o$ denotes the \emph{observed} set of links, which is a subset $E^o \subseteq E^*$ of the complete set of true links $E^* \subseteq V \times V$. Here, $E^*$ encompasses not only the observed links but also potential links that are currently absent or may form in the future within the graph $G$. For any node $v \in V$, $\mathcal{N}(v)=\{u| (u,v) \in E^o \}$ denotes the neighbors of node $v$. The set of $k$-hop simple paths from node $u$ to $v$ is denoted as $\pi_k(u,v) = \{(v_1, v_2, \dots, v_k) | v_1 = u, v_k = v \text{ and } (v_i, v_{i+1}) \in E^o \text{ for } i \in \{1,\dots, k-1\}\}$. Note that paths only contain distinct nodes. We denote the shortest-path between a node pair $(u,v)$ as $\text{SP}(u,v)$.


The objective of LP tasks is to identify the set of unobserved true links $E^u \subseteq E^* \backslash E^o$ within a given graph $G$. This task diverges from typical binary classification problems, as the potential candidates for $E^u $ are predetermined: they consist of all node pairs not already included in the observed links $ V \times V \backslash E^o $. In practical terms, ``identifying'' $E^u$ equates to ranking these unobserved true links higher than false links based on their link features~\cite{yang_evaluating_2015,hu_open_2021}. This ranking process is defined by what we term \emph{connectivity patterns}\footnote{We have an in-depth analysis on how connectivity patterns differ from and relate to graph distributions in Appendix~\ref{app:pattern}, where we illustrate that graphs with different underlying distribution could have the shared connectivity pattern.}:
\begin{definition}
\label{def:connectivity}
\textbf{Connectivity pattern} is an ordered sequence of events $\omega=[A_1,A_2,\dots]$ such that $p(y=1|A_i) \geq p(y=1|A_j)$ for any $i < j$.
\end{definition}
Here, an event $ A $ refers to a specific set of conditions met by the link features of a node pair. In LP tasks, connectivity patterns may be determined by human experts using heuristic methods or by training parametric link predictors. For example, in social networks, a simple connectivity pattern might be $ \omega=[CN(u,v)\geq1, CN(u,v)=0] $, suggesting that pairs of users with common friends are more likely to connect than those without any.

The ability to transfer a connectivity pattern from one graph to another suggests the potential for LP models to be applicable to new, previously unseen graphs. However, a mismatch in the ranking of connectivity patterns between the training and target graphs could lead to inaccuracies, since the model can assign higher scores to unlikely links and lower scores to likely ones.


\begin{figure}[t]
    \centering
\begin{subfigure}[t]{0.4 \linewidth}
    \includegraphics[width=\linewidth]{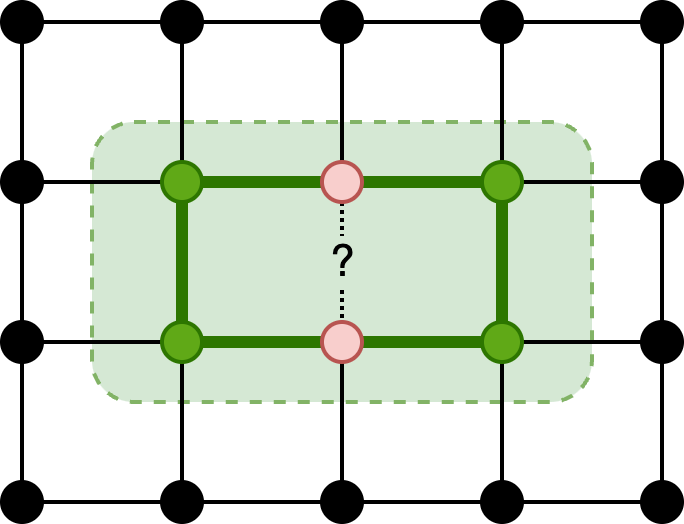}
    \vspace{-6mm}
    \subcaption{\label{fig:grid}}
    
\end{subfigure}
\hspace{5mm}
\begin{subfigure}[t]{0.4 \linewidth}
    \centering
    \includegraphics[width=\linewidth]{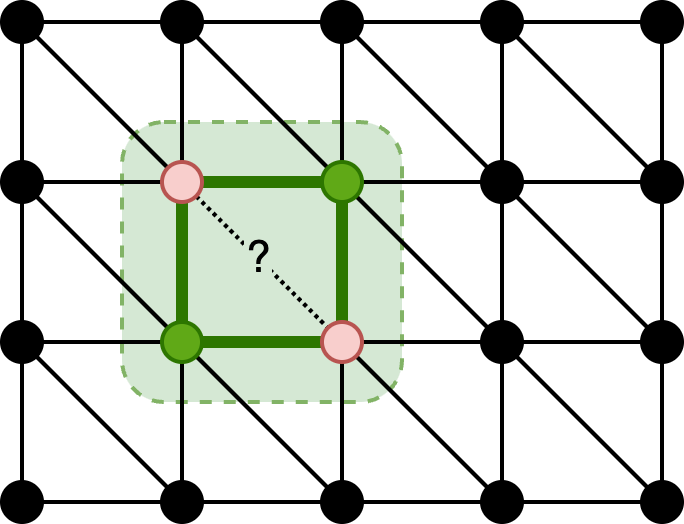}
    \vspace{-6mm}
    \subcaption{\label{fig:tri}}
    
\end{subfigure}
\vspace{-3mm}
\caption{Two synthetic graphs with different connectivity patterns: (a) \textbf{Grid} lattice graph; (b) \textbf{Triangular} lattice graph.}\label{fig:syn}
\vspace{-6mm}
\end{figure}

Next, we demonstrate that even structurally similar synthetic graphs can exhibit different connectivity patterns. We begin by considering two types of lattice graphs: a \textbf{Grid} graph, similar to a chessboard, where nodes are evenly spaced on a 2D grid, each connected to its four nearest neighbors; and a \textbf{Triangular} graph, derived from the Grid by adding one diagonal edge within each square unit. Despite their structural similarities, these graphs, Grid and Triangular, display divergent connectivity patterns:
\begin{theorem}
\label{thm:conflict}
Define $A_2=|\pi_2(u,v)|\geq1$ and $A_3=|\pi_3(u,v)|\geq1$ as elements of $ \omega $. The connectivity patterns on Grid and Triangular graphs are distinct. Specifically:\\
(i) On Grid: $ \omega=[A_3,A_2] $;
(ii) On Triangular: $ \omega=[A_2,A_3] $.
\end{theorem}
The proof is in Appendix~\ref{thm:conflict}. In essence, in Triangular graphs, node pairs two hops away are more likely to form a link compared to those three hops away. Conversely, in Grid graphs, despite their structural similarity to Triangular graphs, node pairs two hops away have no likelihood of linking.

This observation of conflicting connectivity patterns across similar graphs underlines the challenges in knowledge transfer for LP tasks. Even slight structural variations in graphs can significantly alter the likelihood of link formation between nodes. Consequently, the task of developing a universal link predictor, capable of adapting to any graph without specific tuning for its connectivity pattern, is a non-trivial endeavor.



\subsection{Contextualizing Link Prediction}

The challenge of conflicting connectivity patterns across different graphs highlights a critical issue: a model trained on one graph may break down when applied to another without accommodating the unique characteristics of the target graph. To mitigate this, we suggest a paradigm where the model dynamically adapts to the target graph by taking into account its specific characteristics.

This adjustment process involves conditioning the model on the target graph's properties, thereby ensuring that the prediction of link formation, $p(1|A)$, is influenced not just by the inherent link features but also by the properties of the target graph. We draw inspiration from the concept of In-context Learning (ICL) in LLMs~\cite{dai_why_2022}, which enables LLMs to solve tasks with a few demonstration examples. We propose the incorporation of the target graph as a contextual element $c$ in the link prediction $p(1|A,c)$. By doing so, the model learns to understand the joint distribution of link features and the graph context, allowing it to adapt to different graphs. In the subsequent section, we will delve into the practical implementation of an LP model equipped with ICL capabilities.




\section{Universal Link Predictor}

\begin{figure*}[t]
    \centering
\begin{subfigure}[t]{0.6 \textwidth}
    \includegraphics[width=\linewidth]{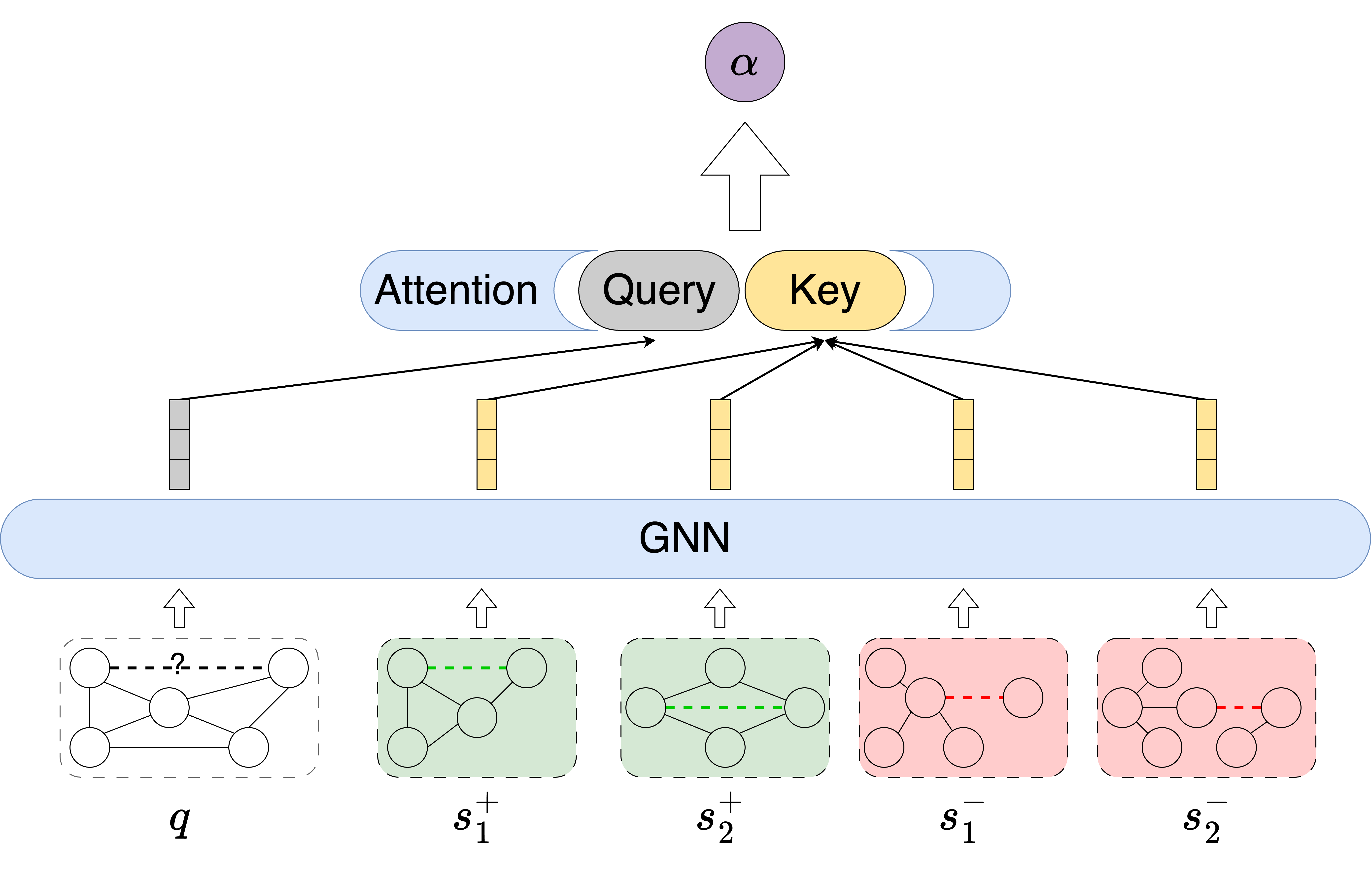}
    \subcaption{\label{fig:encode}}
    
\end{subfigure}%
\hspace{5mm}
\begin{subfigure}[t]{0.3 \textwidth}
    \centering
    \includegraphics[width=\linewidth]{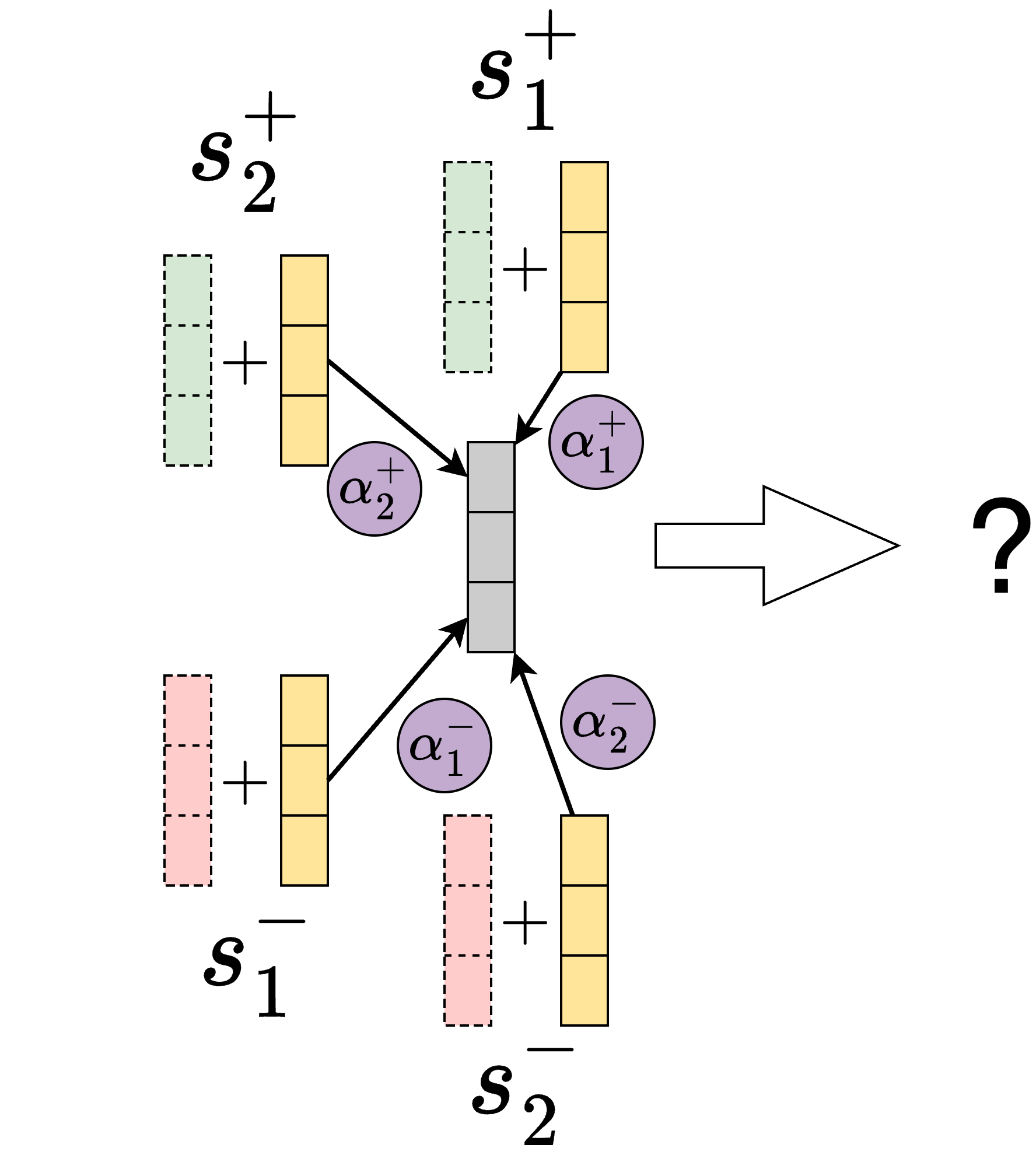}
    \subcaption{\label{fig:agg}}
    
\end{subfigure}
\caption{Overview of the Universal Link Predictor framework. (a) For predicting a query link $ q $, we initially sample positive ($ s^+ $) and negative ($ s^- $) in-context links from the target graph. Both the query link and these in-context links are independently processed through a shared subgraph GNN encoder. An attention mechanism then calculates scores based on the similarity between the query link and the in-context links. (b) The final representation of the query link, contextualized by the target graph, is obtained through a weighted summation, which combines the representations of the in-context links with their respective labels.}\label{fig:framework}
\end{figure*}

This section outlines our proposed \our, designed for effective application to unseen datasets. \our operates by first sampling a set of in-context links from the target graph, which are then independently encoded alongside the target link using a shared GNN encoder. An attention mechanism is employed to merge the representations of these in-context links in relation to their interaction with the target link, forming a composite representation for the final prediction. The overall framework is in Figure~\ref{fig:framework}. \mao{add description on the overall pipeline in combine with graph}

\mao{Most of my papers do not have the methodology section. I am not so good at this part. You can mention which part needs the proofread.}
\subsection{Query and in-context links}
For a given target link $ q \in V \times V $ in graph $ G $, we define it as the \emph{query} link. To predict this link based on the contextual information of $ G $, we start by sampling a set of \emph{in-context} links from $ G $. Specifically, we select $ k $ node pairs $ S^+ \subseteq E^o $ as positive examples, where $ S^+ = \{s_1^+, s_2^+, \dots, s_k^+\} $. These pairs have existing links between them in $ G $. Similarly, we gather negative examples $ S^- = \{s_1^-, s_2^-, \dots, s_k^-\} \subseteq V \times V \backslash E^o $, comprising $ k $ node pairs without a link. The combined set $ S^+ \cup S^- $ approximates the overall properties of $ G $ and provides a context $ c $ for the model to perform LP ($ p(1|A,c) $) using both link features and graph context.

Once we get the query link and the in-context links, we need to obtain the structural representation for them. We start by extracting the ego-subgraph for each of them. An ego-subgraph $\mathbb{G}((u,v),r,G)$ for a node pair $(u,v)$ is a subgraph induced by all the $r$-hop neighboring nodes of the nodes $u$ and $v$ on the graph $G$:
\begin{align*}
    \mathbb{G}((u,v),r,G) = (V_s,E_s),
\end{align*}
where $V_s = \{i|\text{SP}(i,u)\leq r \text{ or } \text{SP}(i,v)\leq r\}$ and $E_s = \{(i,j)\in E^o|i,j\in V_s\}$. For simplicity, we denote such an ego-subgraph as $\mathbb{G}(e)$ for the node pair $e=(u,v)$ when there is no ambiguity. The ego-subgraphs for the query link and the in-context links are $\{\mathbb{G}(e)| e \in \{q\} \bigcup S^+ \bigcup S^- \}$.

Utilizing ego-subgraphs to represent links offers several key advantages over using either individual node pairs or the entire graph. Firstly, an ego-subgraph provides a richer structural context than a mere node pair, encapsulating the local neighborhood structure around the link in question. This approach allows for a more detailed and informative representation of the link's local structures. Secondly, ego-subgraphs serve as an effective and computationally efficient approximation of global link features~\cite{zhang_link_2018}. This is advantageous over the resource-intensive process of encoding the entire graph. Lastly, representations at the subgraph level are inherently more expressive compared to node-level representations~\cite{frasca_understanding_2022}. This enhanced expressiveness is crucial for capturing the structures of links and performing accurate LPs.

\subsection{Encoding ego-subgraphs}
The ego-subgraphs within the set $ \{\mathbb{G}(e)| e \in \{q\} \cup S^+ \cup S^- \} $ vary in size, requiring a uniform approach to representation. We employ GNNs to encode these subgraphs into a consistent latent space.

In the absence of node features in non-attributed graphs, typical GNNs~\cite{kipf_semi-supervised_2017,hamilton_inductive_2018,xu_how_2018} require initial input vectors for each node. Common methods like assigning identical or random vectors meet this requirement but lack expressiveness about graph structures~\cite{li_distance_2020,zhang_labeling_2021}. To address this, we utilize the \emph{labeling trick} technique, assigning each node $ i $ in $ \mathbb{G}(e) $ a positional encoding based on its relative position to the target link $ e=(u,v) $. 

Our approach, \emph{DRNL+}, is a variant of Double Radius Node Labeling (DRNL)~\cite{zhang_link_2018} and Distance Encoding (DE)~\cite{li_distance_2020}. In DRNL, nodes $ i $ in $ \mathbb{G}(e) $ are assigned integer labels as follows:
\begin{align*}
\text{DRNL}(i,(u,v)) = 1 + \text{min}(d_u, d_v) + (d / 2)[(d / 2) + (d \% 2) - 1],
\end{align*}
where $ d_u := \text{SP}(i,u) $, $ d_v := \text{SP}(i,v) $, and $ d := d_u + d_v $. However, DRNL doesn't distinguish nodes reachable to only one of the target nodes. Thus, \emph{DRNL+} enhances this by using DE to assign a tuple of integers:
\begin{align}\label{eq:drnl+}
    \text{DRNL+}(i,(u,v)) = \left\{\begin{alignedat}{2}
    & (0,d_u), &  \text{if } d_v=\infty\\
    & (0,d_v), &  \text{if } d_u=\infty\\
    & (\text{DRNL}(i,(u,v)),0), &  \text{otherwise}
  \end{alignedat}\right.
\end{align}

After the labeling trick indicates relative positions, we apply the SAGE~\cite{hamilton_inductive_2018} with mean aggregation to update node representations. The final subgraph representation, $h_e \in \mathbb{R}^F$ for each link $e \in \{q\} \bigcup S^+ \bigcup S^-$, is derived by average pooling the representations of all nodes in $ \mathbb{G}(e) $. We find that mean aggregation and pooling work best for such a universal link predictor, hypothesizing that this approach better accommodates varying graph sizes and node degrees, thereby enhancing model generalizability.

\subsection{Link prediction with context}

\mao{I think the current definition on the context is not clear. In nlp, it should be (x1,y1, x2, y2, ..., xn, yn)}
Once the ego-subgraphs are encoded into latent space, we utilize these representations $ \{h_e|e \in \{q\} \cup S^+ \cup S^- \} $ to parameterize our link predictor $ p(1|A,c) $ via an attention mechanism~\cite{vaswani_attention_2017}.

The attention scores $ a $ between the query link representation $ h_q $ and each in-context link representation $ h_s $ for $ s \in S^+ \cup S^- $ are calculated using additive attention~\cite{niu_review_2021,brody_how_2022}:
\begin{align}
    a_s=
    p^{\top}
	\mathrm{LeakyReLU}
	\left(
		\mW_k \cdot \left[h_{q} \| h_{s}\right] 
	\right),
\end{align}
where $ p^{\top} \in \mathbb{R}^{F^\prime} $ is a learnable vector and $ \mW_k \in \mathbb{R}^{F^\prime \times 2F} $ is a projection matrix. The concatenation operation is denoted by $ \| $. The normalized attention scores $ \alpha $ are obtained as follows:
\begin{equation}
    \alpha_{s} =
	\softmax\left(
	a_s\right) =
	\frac{\mathrm{exp}\left(a_s\right)}{\sum\nolimits_{e\in S^+ \bigcup S^-} \mathrm{exp}\left(a_e\right)}.
\end{equation}
We denote the attention score between the query link and a positive in-context link $ s^+ \in S^+ $ as $ \alpha^+ $, and with a negative in-context link as $ \alpha^- $. Like in Transformer and GAT models~\cite{velickovic_graph_2018,brody_how_2022}, multi-head attention can also be employed to capture diverse interactions between graph structures.

\paragraph{Remark.} The attention scores are pivotal for shaping the query link's representation in the context of the target graph. We intentionally exclude label information from the attention score computation to avoid biasing the model towards easy predictions during training. This approach aligns with an ``unsupervised'' learning strategy, as opposed to a ``supervised'' one, where label information might lead the model to rely excessively on seen patterns, thus turning the attention mechanism into a \emph{de facto} classifier. This could hinder the model's ability to generalize and adapt across varying graph structures, increasing the risk of overfitting to specific connectivity patterns not applicable to new, unseen graphs. Our empirical findings support this methodology, demonstrating that keeping the attention computation label-free significantly boosts the model's generalizability.

After the normalized attention scores $ \alpha $ are determined, we compute the final representation for the query link $ q $. This is achieved by applying a weighted sum to the representations of the in-context links, using the attention scores as weights. Additionally, we integrate label information into the in-context links' representations by adding corresponding learnable vectors. Formally, the final representation is calculated as follows:
\begin{equation}
    \Tilde{h}_q = \sum\nolimits_{s \in S^+} \alpha^+_{s} \mW_v \left( h_{s} + l^+ \right) + \sum\nolimits_{s \in S^-} \alpha^-_{s} \mW_v \left( h_{s} + l^- \right),
\end{equation}
where $ l^+,l^- \in \mathbb{R}^{F^\prime} $ are learnable vectors for labels, and $ \mW_v \in \mathbb{R}^{F^\prime \times F} $ is a value projection matrix. The representation $ \Tilde{h}_q $ encapsulates both the link features of the query link $ q $ and an estimation of the target graph $ G $, and is then input into an MLP classifier to produce the link prediction result:
\begin{equation}
p(1|A,c) = \sigma\left(\text{MLP}\left(\Tilde{h}_q\right)\right),
\end{equation}
where $ \sigma\left(\cdot\right) $ denotes a sigmoid function.

\subsection{Pretraining objective}

The pretraining objective for \our focuses on predicting the query link $ q $ based on its own features and the context of the graph it is part of. We align this objective with standard binary classification as seen in typical parametric link prediction algorithms~\cite{zhang_link_2018,chamberlain_graph_2022,dong_pure_2023}. In this setting, the classification label $ y_e $ for an edge $ e $ is set to 1 if $ e $ is among the observed links $ E^o $; otherwise, $ y_e $ is 0. Additionally, we consider a set of pretrain graphs $ \mathcal{G} $, with each graph $ G $ being a member of this set. The overall pretraining loss is then defined as:
\begin{equation}
    \mathcal{L} = \E_{G \in \mathcal{G},e \in V \times V}{ \text{BCE}\left ( \text{MLP}\left(\Tilde{h}_e\right),y_e \right )}.
\end{equation}
This loss function is employed across multiple graphs, allowing \our to learn a generalizable pattern for link prediction across various graph structures.

\section{Related work} 
\mao{Can we move the related work to an early stage, maybe between the observation and methods} 

\paragraph{Link prediction.}
Traditional LP methods are handcrafted heuristics designed by observing the connectivity pattern in real-world data. They leverage either the link's local~\cite{liben-nowell_link_2003,adamic_friends_2003,zhou_predicting_2009,barabasi_emergence_1999} or global information~\cite{katz_new_1953,page_pagerank_1999} to infer the missing links in the graph. OLP~\cite{ghasemian_stacking_2020} stacks the heuristics link predictors as a feature vector and fits a random forest as the classifier. WLNM~\cite{zhang_weisfeiler-lehman_2017} is one of the pioneers in training a neural network as a link predictor. GAE~\cite{kipf_variational_2016}, as the first GNN4LP,  utilizes GNNs to encode the graph structure into node representation and perform the link prediction task. SEAL~\cite{zhang_link_2018,zhang_labeling_2021} points out that a link-level representation is necessary for a successful LP method and proposes the labeling trick to enable GNNs to learn the joint structural representation. ELPH~\cite{chamberlain_graph_2022}, NCNC~\cite{wang_neural_2023}, and MPLP~\cite{dong_pure_2023} further improve the scalability of GNN4LP and achieve the state-of-the-arts on various graph benchmarks.

\paragraph{In-context Learning.}
The remarkable efficacy of LLMs across a broad spectrum of language tasks is significantly attributed to their adeptness in ICL~\cite{brown_language_2020}. This capability allows LLMs to generalize to new tasks by leveraging demonstration examples, effectively learning the required skills on the fly. ~\citet{irie_dual_2022} delves into the equivalence between conventional model training and the application of attention mechanisms to training samples during inference, suggesting an underlying mechanism of ICL. Further exploration by \citet{dai_why_2022} posits that ICL facilitates an implicit optimization process guided by in-context examples. While the concept of ICL has been primarily associated with LLMs, Prodigy~\cite{huang_prodigy_2023} represents an initial attempt to adapt ICL for GNN-based models. Their approach, however, is somewhat constrained by the overlap in pretrain and test datasets, which raises questions about the method's transferability across distinct graph domains. 




\section{Experiments}
In this section, we conduct extensive experiments to assess the performance of \our on new unseen datasets.

\subsection{Experimental setup}
\mao{If the space is not enough, I think we can remove this section title}
\paragraph{Benchmark datasets.} The foundation for our model's training is a collection of graph datasets spanning a variety of domains. Following~\cite{mao_revisiting_2023}, we have carefully selected \mao{diverse} graph data from fields such as biology~\cite{von_mering_comparative_2002,zhang_beyond_2018,watts_collective_1998}, transport~\cite{watts_collective_1998,batagelj_pajek_2006}, web~\cite{ackland_mapping_2005,spring_measuring_2002,adamic_political_2005}, academia collaboration~\cite{shchur_pitfalls_2019,newman_finding_2006}, citation~\cite{yang_revisiting_2016}, and social networks~\cite{rozemberczki_multi-scale_2021}. This diverse selection ensures that we can pretrain and evaluate the LP model based on a wide array of connectivity patterns. The details of the curated graph datasets can be found in Table~\ref{tab:stats} in Appendix.


\paragraph{Baseline Methods.} We compare \our with both heuristic and GNN-based parametric link predictors. Heuristic methods include Common Neighbor (CN)~\cite{liben-nowell_link_2003}, Adamic-Adar index (AA)~\cite{adamic_friends_2003}, Resource Allocation (RA)~\cite{zhou_predicting_2009}, Preferential Attachment (PA)~\cite{barabasi_emergence_1999}, Shortest-Path (SP), and Katz index (Katz)~\cite{katz_new_1953}. GNN-based methods include GAE~\cite{kipf_variational_2016}, SEAL~\cite{zhang_link_2018}, ELPH~\cite{chamberlain_graph_2022}, NCNC~\cite{wang_neural_2023}, and MPLP~\cite{dong_pure_2023}.  For GAE and NCNC, which require initial node features, we use a $32$-dimensional all-one vector. All other methods can handle non-attributed graphs directly.

\paragraph{Evaluation of \our.} To evaluate \our's effectiveness on unseen datasets, we divide our graph data into non-overlapping pretrain and testing sets (see Table~\ref{tab:stats}) and pretrain one single model on the combined pretrain datasets. During pretraining, we dynamically sample $40$ positive and negative links as in-context links $S^+ \cup S^-$ for each query link from the corresponding pretrain dataset. For evaluation, each test dataset is split into 70\%/10\%/20\% for training/validation/testing. The training set here forms the observed links $E^o$, while validation and test sets represent unobserved links $E^u$. During the inference, we sample $200$ positive and negative links as in-context links per test dataset. We report Hits@50~\cite{hu_open_2021} as the evaluation metric for LP. More details about the pretraining of \our can be found in Appendix~\ref{app:pretrain}. 
\mao{I think since we only pre-train on one dataset we need to emphasize which dataset we utilize for pre-train and how it divergers from other dataset. This is an important experimental setting}

\paragraph{Evaluation of Baselines.} Baseline models follow similar evaluation procedures, with adaptations for transfer learning capabilities. We employ two settings: (1) \textbf{Pretrain Only}, where models are trained on combined pretrain datasets \mao{Why here is combined pretrain datasets} and then tested on each test dataset, and (2) \textbf{Pretrain \& Finetune}, where after pretraining, models are additionally finetuned on each test dataset with $200$ sampled positive and negative links for training.

\subsection{Primary results}

\begin{table*}[h]
    \centering
    \caption{Link prediction results on test datasets evaluated by Hits@50. The format is average score ± standard deviation. The top three models are colored by \textbf{\textcolor{red}{First}}, \textbf{\textcolor{blue}{Second}}, \textbf{\textcolor{violet}{Third}}.}\label{tab:real}
    \resizebox{\textwidth}{!}{%
    \begin{tabular}{lcccccccc}
    \toprule
            &\multicolumn{1}{c}{Biology}&\multicolumn{1}{c}{Transport}&\multicolumn{1}{c}{Web}&\multicolumn{2}{c}{Collaboration}&\multicolumn{1}{c}{Citation}&\multicolumn{1}{c}{Social}\\ \cmidrule(r{0.5em}){2-2} \cmidrule(r{0.5em}){3-3} \cmidrule(r{0.5em}){4-4} \cmidrule(r{0.5em}){5-6} \cmidrule(r{0.5em}){7-7} \cmidrule(r{0.5em}){8-8} 
            
        & \textbf{C.ele} & \textbf{USAir} & \textbf{PB} & \textbf{NS} & \textbf{CS} & \textbf{Cora} & \textbf{Facebook} & \textbf{Ave. Rank}\\
    
        \midrule
        \rowcolor{Gray}
        \multicolumn{9}{c}{Heuristics} \\
        \textbf{CN} & $46.88 {\scriptstyle \pm 12.28}$ & $82.75 {\scriptstyle \pm 1.54}$ & $41.15 {\scriptstyle \pm 3.77}$ & $74.03 {\scriptstyle \pm 1.59}$ & $56.84 {\scriptstyle \pm 15.56}$ & $33.85 {\scriptstyle \pm 0.93}$ & $58.70 {\scriptstyle \pm 0.35}$ & 11.00 \\
    \textbf{AA} & $61.07 {\scriptstyle \pm 5.16}$ & \thirdt{86.96}{2.24} & $44.12 {\scriptstyle \pm 3.36}$ & $74.03 {\scriptstyle \pm 1.59}$ & \thirdt{68.22}{1.08} & $33.85 {\scriptstyle \pm 0.93}$ & \secondt{67.80}{2.12} & 5.71 \\
    \textbf{RA} & $62.80 {\scriptstyle \pm 4.84}$ & \secondt{87.27}{1.89} & $43.72 {\scriptstyle \pm 2.86}$ & $74.03 {\scriptstyle \pm 1.59}$ & $68.21 {\scriptstyle \pm 1.08}$ & $33.85 {\scriptstyle \pm 0.93}$ & \firstt{68.84}{2.03} & 5.57 \\
    \textbf{PA} & $43.85 {\scriptstyle \pm 4.12}$ & $77.69 {\scriptstyle \pm 2.29}$ & $28.93 {\scriptstyle \pm 1.91}$ & $35.35 {\scriptstyle \pm 3.01}$ & $6.49 {\scriptstyle \pm 0.61}$ & $22.09 {\scriptstyle \pm 1.52}$ & $12.95 {\scriptstyle \pm 0.63}$ & 13.71 \\
    \textbf{SP} & $0.00 {\scriptstyle \pm 0.00}$ & $0.00 {\scriptstyle \pm 0.00}$ & $0.00 {\scriptstyle \pm 0.00}$ & $80.00 {\scriptstyle \pm 1.11}$ & $41.34 {\scriptstyle \pm 35.58}$ & $52.97 {\scriptstyle \pm 1.53}$ & $0.00 {\scriptstyle \pm 0.00}$ & 13.14 \\
    \textbf{Katz} & $58.86 {\scriptstyle \pm 6.48}$ & $84.64 {\scriptstyle \pm 1.86}$ & $44.36 {\scriptstyle \pm 3.65}$ & $78.96 {\scriptstyle \pm 1.35}$ & $66.32 {\scriptstyle \pm 5.59}$ & $52.97 {\scriptstyle \pm 1.53}$ & $60.79 {\scriptstyle \pm 0.60}$ & 6.86 \\
    
        \midrule
        \rowcolor{Gray}
        \multicolumn{9}{c}{Pretrain Only} \\
        \textbf{SEAL} & $61.28 {\scriptstyle \pm 3.76}$ & $86.00 {\scriptstyle \pm 1.56}$ & $45.44 {\scriptstyle \pm 2.68}$ & $84.07 {\scriptstyle \pm 1.96}$ & $62.82 {\scriptstyle \pm 1.62}$ & $56.21 {\scriptstyle \pm 2.24}$ & $54.57 {\scriptstyle \pm 1.48}$ & 5.57 \\
    \textbf{GAE} & $44.71 {\scriptstyle \pm 3.39}$ & $76.12 {\scriptstyle \pm 2.27}$ & $27.56 {\scriptstyle \pm 2.34}$ & $15.20 {\scriptstyle \pm 2.14}$ & $5.08 {\scriptstyle \pm 0.48}$ & $24.22 {\scriptstyle \pm 1.53}$ & $6.65 {\scriptstyle \pm 0.57}$ & 14.14 \\
    \textbf{ELPH} & $59.23 {\scriptstyle \pm 4.50}$ & $84.42 {\scriptstyle \pm 2.22}$ & $43.69 {\scriptstyle \pm 2.90}$ & $84.27 {\scriptstyle \pm 1.43}$ & \secondt{70.69}{3.63} & $56.91 {\scriptstyle \pm 1.43}$ & $61.80 {\scriptstyle \pm 2.46}$ & 5.57 \\
    \textbf{NCNC} & $48.07 {\scriptstyle \pm 4.79}$ & $75.44 {\scriptstyle \pm 4.22}$ & $25.66 {\scriptstyle \pm 1.42}$ & $80.07 {\scriptstyle \pm 1.43}$ & $34.27 {\scriptstyle \pm 2.28}$ & $52.51 {\scriptstyle \pm 2.54}$ & $19.28 {\scriptstyle \pm 1.58}$ & 11.57 \\
    \textbf{MPLP} & $56.74 {\scriptstyle \pm 5.31}$ & $82.94 {\scriptstyle \pm 2.30}$ & \thirdt{47.78}{2.55} & $80.33 {\scriptstyle \pm 1.54}$ & $24.26 {\scriptstyle \pm 1.28}$ & $46.71 {\scriptstyle \pm 2.25}$ & $48.06 {\scriptstyle \pm 2.09}$ & 8.86 \\
    
        \midrule
        \rowcolor{Gray}
        \multicolumn{9}{c}{Pretrain \& Finetune} \\
        \textbf{SEAL} & \thirdt{64.45}{4.14} & \firstt{88.49}{2.16} & $47.78 {\scriptstyle \pm 3.32}$ & $84.84 {\scriptstyle \pm 2.32}$ & $61.54 {\scriptstyle \pm 3.09}$ & \firstt{62.19}{3.27} & $58.70 {\scriptstyle \pm 2.78}$ & \secondts{3.14} \\
    \textbf{GAE} & $44.71 {\scriptstyle \pm 4.07}$ & $74.47 {\scriptstyle \pm 2.96}$ & $25.92 {\scriptstyle \pm 2.64}$ & $18.34 {\scriptstyle \pm 2.27}$ & $4.95 {\scriptstyle \pm 0.44}$ & $25.31 {\scriptstyle \pm 1.48}$ & $6.11 {\scriptstyle \pm 0.39}$ & 14.71 \\
    \textbf{ELPH} & $60.51 {\scriptstyle \pm 5.72}$ & $84.52 {\scriptstyle \pm 2.08}$ & $43.58 {\scriptstyle \pm 3.48}$ & \thirdt{86.08}{0.69} & \firstt{71.10}{3.48} & $57.18 {\scriptstyle \pm 1.89}$ & $63.31 {\scriptstyle \pm 3.63}$ & 4.71 \\
    \textbf{NCNC} & \secondt{64.45}{5.10} & $85.85 {\scriptstyle \pm 2.46}$ & $47.75 {\scriptstyle \pm 6.90}$ & \secondt{88.25}{2.25} & $58.75 {\scriptstyle \pm 5.74}$ & \secondt{60.00}{2.50} & $59.32 {\scriptstyle \pm 6.93}$ & \thirdts{3.71} \\
    \textbf{MPLP} & $62.56 {\scriptstyle \pm 4.79}$ & $85.08 {\scriptstyle \pm 1.54}$ & \secondt{48.01}{2.94} & $80.16 {\scriptstyle \pm 1.18}$ & $50.35 {\scriptstyle \pm 1.01}$ & $56.02 {\scriptstyle \pm 2.09}$ & $56.72 {\scriptstyle \pm 1.20}$ & 6.14 \\
    
        \midrule
        \rowcolor{Gray}
        \multicolumn{9}{c}{Ours} \\
        \textbf{UniLP} & \firstt{65.20}{4.40} & $85.98 {\scriptstyle \pm 2.00}$ & \firstt{48.14}{2.99} & \firstt{89.09}{2.05} & $64.59 {\scriptstyle \pm 2.65}$ & \thirdt{57.50}{2.40} & \thirdt{65.49}{2.05} & \firstts{1.86} \\
    \bottomrule
    \end{tabular}
}
\end{table*}

Table~\ref{tab:real} presents the performance of \our on various unseen graph datasets. The results demonstrate that \our outperforms both traditional heuristic methods and standard GNN-based LP models that are pretrained without specific adaptation, showing significant improvements in 4 out of 7 the benchmark datasets. This performance enhancement suggests that tailoring the LP model to individual graphs can markedly increase its transfer learning capabilities.

\mao{Can we claim the finetuning is time consuming, and requires hyperparameter tuning}
Moreover, \our achieves comparable or even superior results to GNN-based LP models that undergo finetuning, \mao{Put the following on the front?} despite not being explicitly trained on the test data. This highlights the effectiveness of the ICL capability in \our, which allows the model to adapt seamlessly to specific graph datasets without the need for additional training. By leveraging in-context links provided during the inference phase, \our can dynamically adjust its knowledge of connectivity patterns, demonstrating its potential to deliver robust performance across a wide range of unseen graph datasets. In addition, the experimental results on the synthetic Triangular/Grid lattice graphs can be found in Table~\ref{tab:syn} in the Appendix.

\subsection{The inner mechanism of \our}
\begin{table*}[h]
    \centering
    \caption{Link prediction results on test datasets evaluated by Hits@50 under context perturbation. This table presents the outcomes of link prediction when the context, i.e., in-context links, is deliberately altered. The aim is to analyze how changes in the context influence the final prediction accuracy.}\label{tab:perturb}
    \resizebox{\textwidth}{!}{%
    \begin{tabular}{lccccccc}
    \toprule
            &\multicolumn{1}{c}{Biology}&\multicolumn{1}{c}{Transport}&\multicolumn{1}{c}{Web}&\multicolumn{2}{c}{Collaboration}&\multicolumn{1}{c}{Citation}&\multicolumn{1}{c}{Social}\\ \cmidrule(r{0.5em}){2-2} \cmidrule(r{0.5em}){3-3} \cmidrule(r{0.5em}){4-4} \cmidrule(r{0.5em}){5-6} \cmidrule(r{0.5em}){7-7} \cmidrule(r{0.5em}){8-8} 
            
        & \textbf{C.ele} & \textbf{USAir} & \textbf{PB} & \textbf{NS} & \textbf{CS} & \textbf{Cora} & \textbf{Facebook}\\\midrule
    \textbf{UniLP-FlipLabel} & $0.61 {\scriptstyle \pm 0.27}$ & $15.81 {\scriptstyle \pm 13.17}$ & $0.03 {\scriptstyle \pm 0.03}$ & $27.97 {\scriptstyle \pm 4.29}$ & $0.60 {\scriptstyle \pm 0.23}$ & $2.03 {\scriptstyle \pm 0.55}$ & $0.32 {\scriptstyle \pm 0.15}$ \\
    \textbf{UniLP-RandomContext} & $52.89 {\scriptstyle \pm 5.90}$ & $81.91 {\scriptstyle \pm 2.14}$ & $47.47 {\scriptstyle \pm 3.05}$ & $85.60 {\scriptstyle \pm 1.23}$ & $47.80 {\scriptstyle \pm 6.48}$ & $37.62 {\scriptstyle \pm 5.64}$ & $22.17 {\scriptstyle \pm 6.55}$ \\ \midrule
    \textbf{UniLP} & $65.20 {\scriptstyle \pm 4.40}$ & $85.98 {\scriptstyle \pm 2.00}$ & $48.14 {\scriptstyle \pm 2.99}$ & $89.09 {\scriptstyle \pm 2.05}$ & $64.59 {\scriptstyle \pm 2.65}$ & $57.50 {\scriptstyle \pm 2.40}$ & $65.49 {\scriptstyle \pm 2.05}$ \\
    \bottomrule
    \end{tabular}
}
\end{table*}
\begin{figure*}[h]
\begin{center}
\centerline{\includegraphics[width=\textwidth]{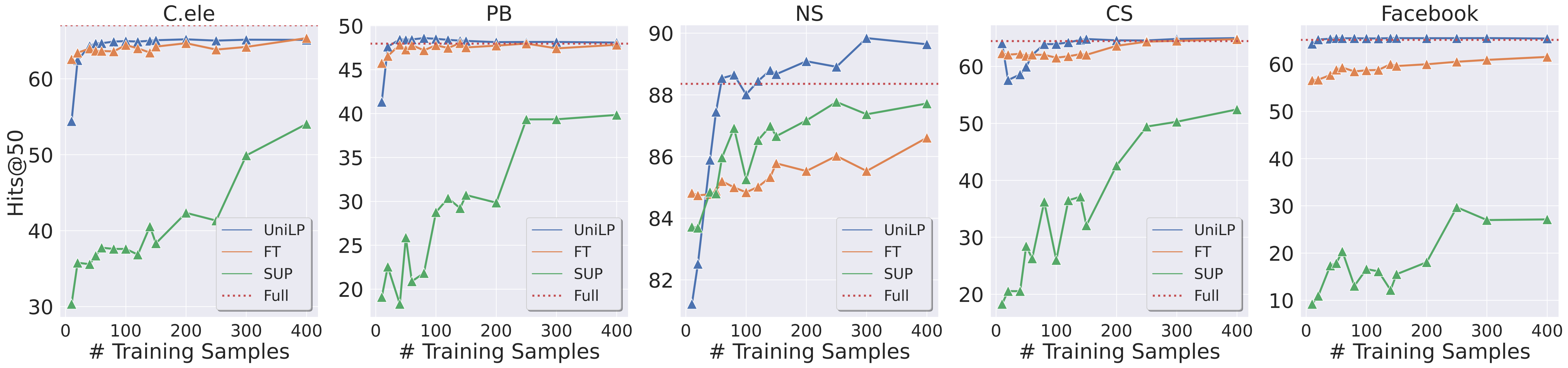}}
\caption{Performance of \our with varying quantities of in-context links.}\label{fig:more_context}
\end{center}
\end{figure*}


We further explore the capability of our proposed model's ICL to facilitate skill learning~\cite{pan_what_2023,mao_data_2024}, enabling the model to acquire new skills not encountered during the pretraining phase, guided by ICL demonstrations. This investigation focuses on the model's performance sensitivity to corrupting in-context links, particularly when these links are presented with incorrect input-label associations. Given that each in-context link consists of an input and its corresponding label, we introduce two perturbation strategies to assess this sensitivity: \textbf{FlipLabel}: we invert the labels of the in-context links, labeling previously positive links as negative and vice versa. \textbf{RandomContext}: Instead of selecting in-context links from the target graph, we randomly sample them from a graph generated using the Stochastic Block Model~\cite{holland_stochastic_1983}. 

The outcomes, as shown in Table~\ref{tab:perturb}, reveal that flipping the labels of in-context links significantly degrades the model's performance, rendering it almost ineffective. This finding underscores the model's utilization of ICL for skill learning, specifically in learning new feature-label mappings within a given context~\cite{wei_larger_2023,min_rethinking_2022}. It highlights the pivotal role of accurate label information in in-context links for the model's effective adaptation to the target graph.

Furthermore, using randomly generated graphs as a source of in-context links also detrimentally affects performance, albeit to varying extents across different datasets. This implies the importance of choosing in-context links that genuinely represent the properties of the target graph. Interestingly, the less severe performance decline in some datasets may indicate their inherent community-based graph structures.

\subsection{Effectiveness of in-context links' size}
This experiment evaluates how varying the quantity of in-context links affects \our's performance during inference. We experiment with different numbers of in-context links, ranging from 10 to 400, sampled from each test graph \mao{What is sampled from each test graph means?}. These links are used as context for the model. Additionally, we utilize SEAL as the base model and assess its performance under finetuning (FT) and supervised training (SUP) from scratch with varying training sample sizes. For comparison, we also include results from training a SEAL model on the full set of target graph data (Full), as detailed in Figure~\ref{fig:more_context}.

The findings reveal a consistent improvement in \our's performance with an increasing number of in-context links. \mao{How about compare with other methods the relatively increasing speed.} This indicates that our method can more effectively capture the target graph's properties with additional context. Notably, on four of the test datasets, \our either matches or exceeds the performance of models trained end-to-end on the entire graph. This suggests that leveraging more pretrpretrain data can be advantageous for LP tasks when properly managed. Furthermore, despite both \our and the finetuned models being pretrained on the same datasets and using the same in-context links, \our occasionally outperforms its finetuned counterparts. This observation suggests that in some cases, utilizing ICL can be a more effective approach for adapting a pretrained model to a specific target dataset compared to finetuning. The trend on the rest of graphs can be found in Figure~\ref{fig:more_context_rest}.

\subsection{Visualization of the link representation}
\begin{figure}[h]
    \centering
\begin{subfigure}[h]{0.5 \linewidth}
    \includegraphics[width=\linewidth]{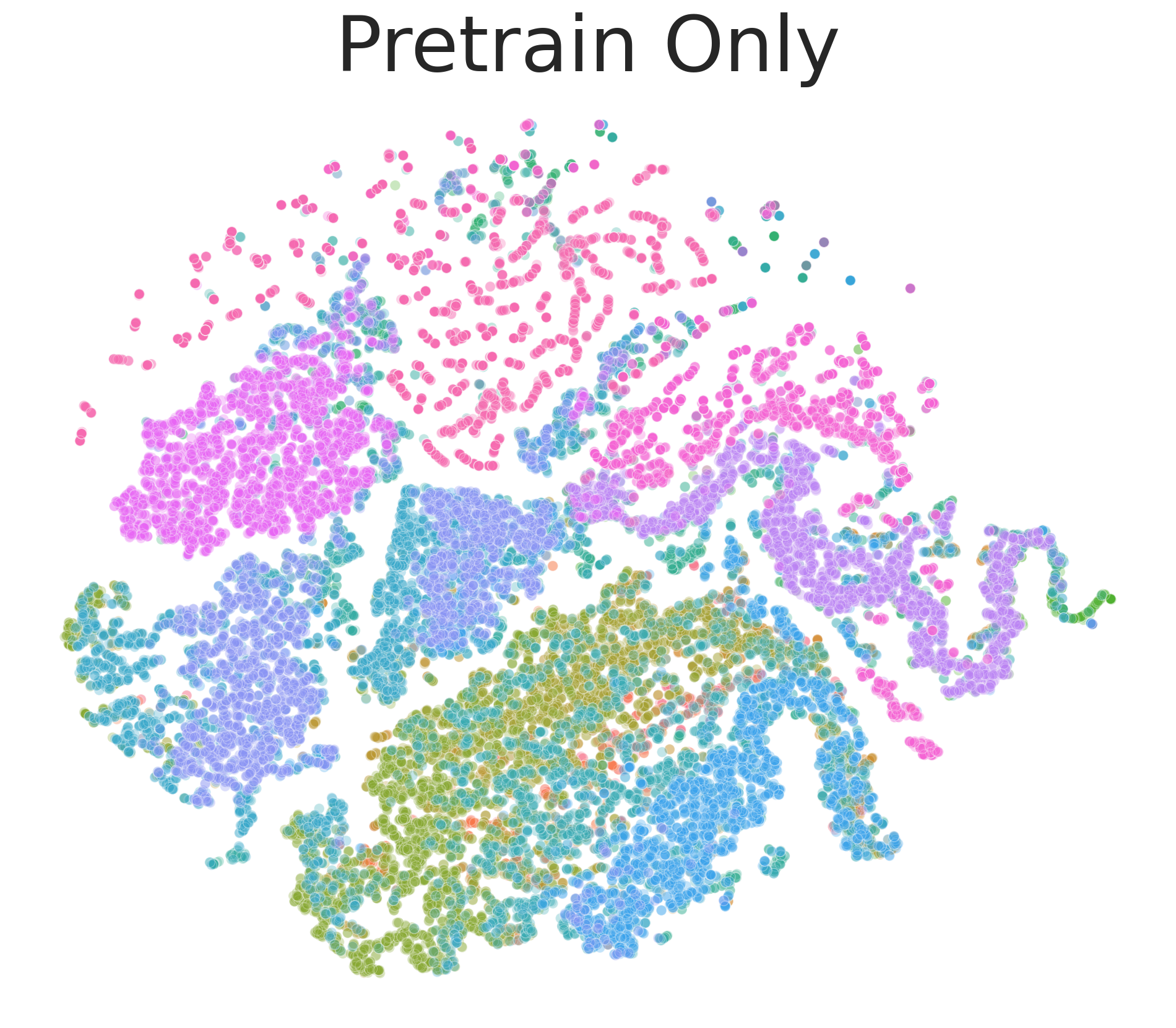}
    \vspace{-6mm}
    \subcaption{\label{fig:pretrain}}
    
\end{subfigure}%
\begin{subfigure}[h]{0.5 \linewidth}
    \centering
    \includegraphics[width=\linewidth]{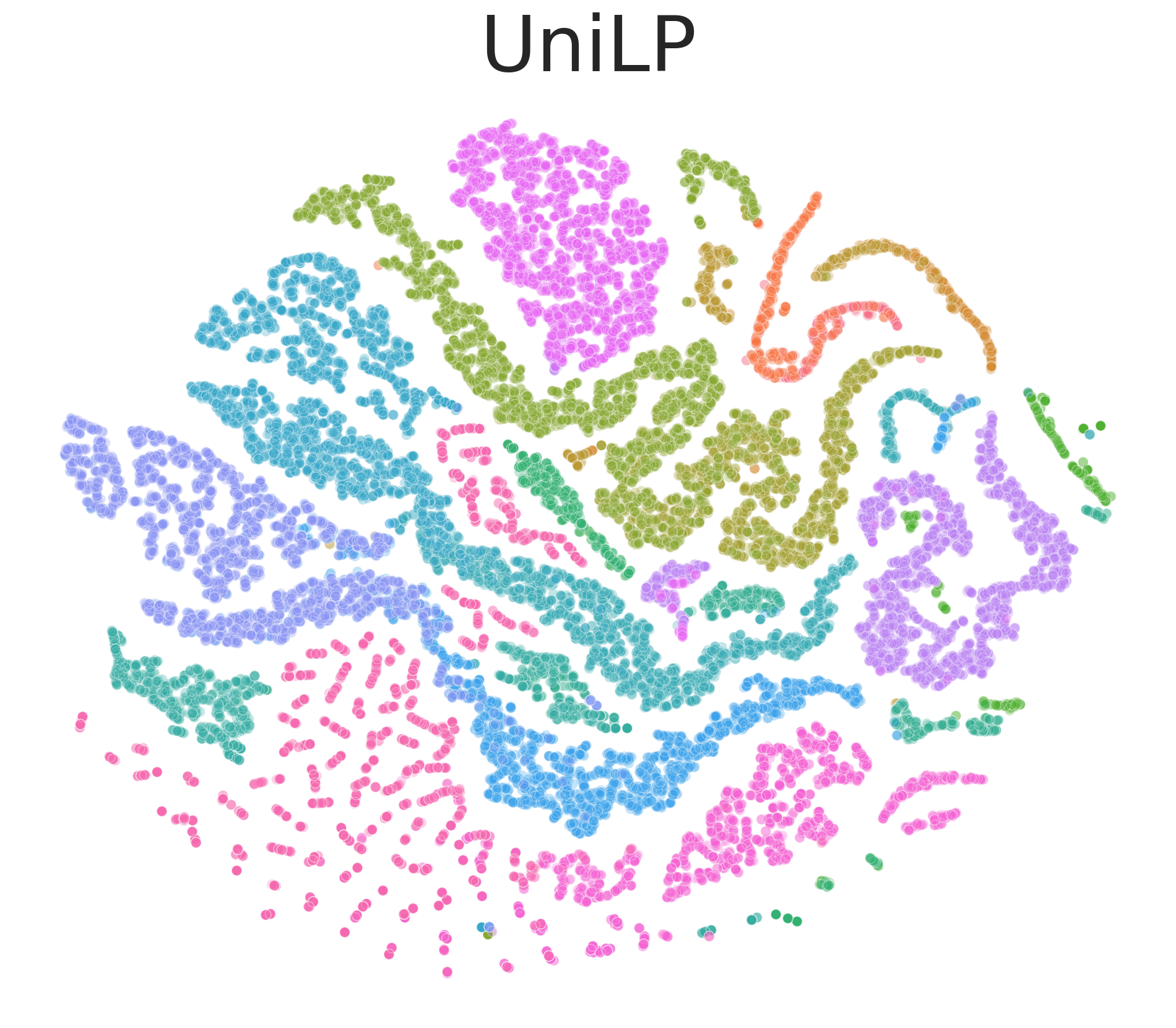}
    \vspace{-6mm}
    \subcaption{\label{fig:icl}}
    
\end{subfigure}
\vspace{-3mm}
\caption{Visualization of the link representation learned from (a) Pretrain Only SEAL; (b) \our. Different colors indicate different test datasets.}\label{fig:tsne}
\vspace{-4mm}
\end{figure}

We conduct a comparative visualization of link representations as learned by a Pretrained Only SEAL model and \our. This comparison is shown in Figure~\ref{fig:tsne}. The results indicate that a naively pretrained model tends to map link representations from various graph datasets into a close subspace, potentially leading to indistinguishable link representations across different graphs, even when these graphs exhibit conflicting connectivity patterns.

In contrast, the link representations generated by \our, which are conditioned on the context of the target graph, demonstrate a distinct separation between different datasets. This separation is indicative of \our's effective ICL capability, which adeptly captures the subtle distributional differences across graphs. By adjusting the link representations based on the context provided by in-context links, \our can effectively address the challenge of conflicting connectivity patterns in diverse graph datasets.
\vspace{-3mm}
\subsection{Additional experiments}
We carry out further experiments to investigate the ICL capabilities of \our: \textbf{Diversifying in-context links selection} (Appendix~\ref{app:diverse}): We explore how varying the selection of in-context links influences \our's LP performance. We find that \our exhibits varying degrees of robustness across different datasets. \textbf{Adjusting positive-to-negative in-context link ratios} (Appendix~\ref{app:ratio}): Instead of employing a balanced ratio of positive and negative in-context links during testing, we experiment with different ratios, including sets composed entirely of positive/negative samples. Surprisingly, we observe exclusive negative in-context links can achieve comparable performance to balanced sets on many datasets.
\section{Conclusion}

In this paper, we introduce the Universal Link Predictor, a novel approach designed to be immediately applicable to any unseen graph dataset without the necessity of training or finetuning. Recognizing the issue of conflicting connectivity patterns among diverse graph datasets, we innovatively employ ICL to dynamically adjust link representations according to the specific properties of the target graph by conditioning on support links as contextual input. Through extensive experimental evaluations, we have demonstrated the effectiveness of our method. Notably, our Universal Link Predictor excels in its ability to adapt seamlessly to new, unseen graphs, surpassing traditional models that require explicit training. This significant advancement presents a promising direction for future research and applications in the field of LP.

\clearpage
\bibliographystyle{ACM-Reference-Format}
\balance
\bibliography{references}

\clearpage
\appendix

\section{Experimental details}

\subsection{Pretrain and test benchmarks}
\begin{table*}[t]
    \centering
    \caption{The pretrain datasets and test benchmarks.}\label{tab:stats}
    \resizebox{\linewidth}{!}{%
    \begin{tabular}{lcc|cccccc}
    \toprule
        \textbf{Dataset} & \textbf{Pretrain} & \textbf{Test} & \textbf{\# Nodes} & \textbf{\# Edges} & \textbf{Avg. node deg.} & \textbf{Std. node deg.} & \textbf{Max. node deg.} & \textbf{Density}\\
    \midrule\rowcolor{Gray}
    \multicolumn{9}{c}{\textbf{Biology}} \\
    Ecoli &\Checkmark & - & 1805 & 29320 & 16.24 & 48.38 & 1030 & 1.8009\%\\
    Yeast &\Checkmark & - & 2375 & 23386 & 9.85 & 15.5 & 118 & 0.8295\%\\
    Celegans &- & \Checkmark & 297 & 4296 & 14.46 & 12.97 & 134 & 9.7734\%\\
    \midrule
    \rowcolor{Gray}
    \multicolumn{9}{c}{\textbf{Transport}} \\
    Power &\Checkmark & - & 4941 & 13188 & 2.67 & 1.79 & 19 & 0.1081\%\\
    USAir &- & \Checkmark & 332 & 4252 & 12.81 & 20.13 & 139 & 7.7385\%\\
    \midrule
    \rowcolor{Gray}
    \multicolumn{9}{c}{\textbf{Web}} \\
    PolBlogs &\Checkmark & - & 1490 & 19025 & 12.77 & 20.73 & 256 & 1.7150\%\\
    Router &\Checkmark & - & 5022 & 12516 & 2.49 & 5.29 & 106 & 0.0993\%\\
    PB &- & \Checkmark & 1222 & 33428 & 27.36 & 38.42 & 351 & 4.4808\%\\
    \midrule
    \rowcolor{Gray}
    \multicolumn{9}{c}{\textbf{Collaboration}} \\
    Physics &\Checkmark & - & 34493 & 495924 & 14.38 & 15.57 & 382 & 0.0834\%\\
    CS &- & \Checkmark & 18333 & 163788 & 8.93 & 9.11 & 136 & 0.0975\%\\
    NS &- & \Checkmark & 1589 & 5484 & 3.45 & 3.47 & 34 & 0.4347\%\\
    \midrule
    \rowcolor{Gray}
    \multicolumn{9}{c}{\textbf{Citation}} \\
    Pubmed &\Checkmark & - & 19717 & 88648 & 4.5 & 7.43 & 171 & 0.0456\%\\
    Citeseer &\Checkmark & - & 3327 & 9104 & 2.74 & 3.38 & 99 & 0.1645\%\\
    Cora &- & \Checkmark & 2708 & 10556 & 3.9 & 5.23 & 168 & 0.2880\%\\
    \midrule
    \rowcolor{Gray}
    \multicolumn{9}{c}{\textbf{Social}} \\
    Twitch &\Checkmark & - & 34118 & 429113 & 12.58 & 35.88 & 1489 & 0.0737\%\\
    Github &\Checkmark & - & 37700 & 289003 & 7.67 & 46.59 & 6809 & 0.0407\%\\
    Facebook &- & \Checkmark & 22470 & 171002 & 7.61 & 15.26 & 472 & 0.0677\%\\
    \bottomrule
    \end{tabular}
}
\end{table*}
Comprehensive details of the curated pretrain and test graph datasets are provided in Table~\ref{tab:stats}. These datasets, selected from various domains and featuring diverse graph statistics, are specifically chosen to ensure that our proposed \our model is exposed to a wide range of LP connectivity patterns. This diversity in training data is crucial for enabling \our to effectively adapt to new, unseen graphs.

\begin{figure*}[h]
\begin{center}
\centerline{\includegraphics[width=\textwidth]{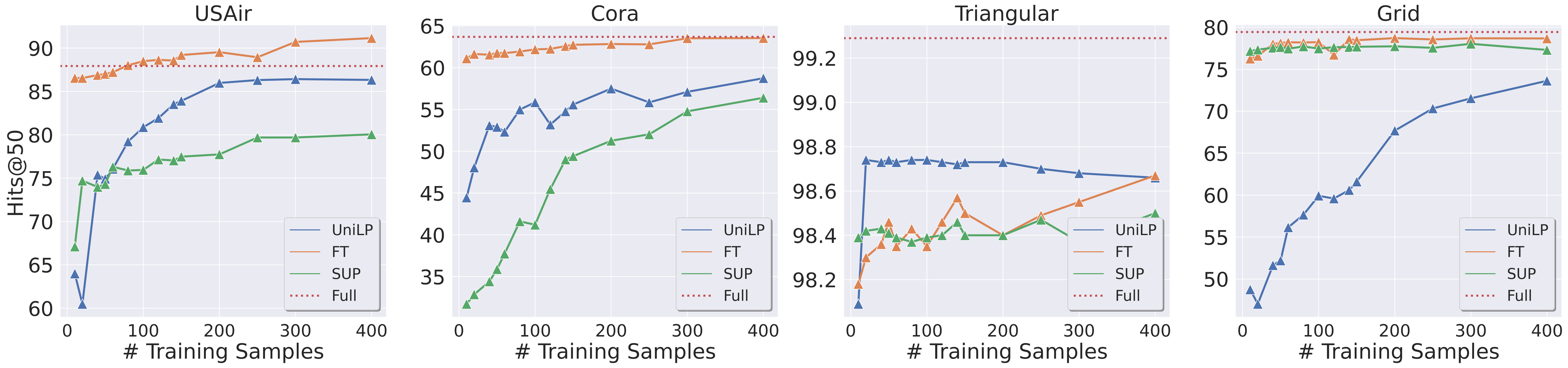}}
\caption{Performance of \our with varying quantities of in-context links on the rest of graph datasets.}\label{fig:more_context_rest}
\end{center}
\end{figure*}

\subsection{Pretraining the Models}
\label{app:pretrain}
We pretrain \our on the datasets listed in Table~\ref{tab:stats}, employing an approach by sampling an equal number of non-connected node pairs ($V \times V \backslash E^o$) as negative samples to match the count of observed links ($|E^o|$) in each graph. The pretraining follows a standard binary classification framework.

For predicting each query link, we sample $40$ positive and negative links as in-context links ($S^+ \cup S^-$) from the respective pretrain dataset. This setup ensures variability: different query links from the same dataset or the same query link across training batches may be paired with different in-context links. However, during testing, the set of in-context links for each test dataset remains constant. This training methodology serves multiple purposes: it enhances \our's generalization capabilities by exposing it to a broad range of in-context links and optimizes GPU memory usage by selecting a manageable yet diverse set of in-context links during pretraining.

The pretraining phase incorporates an early stopping criterion based on performance across a merged validation set, which comprises $200$ links from the validation set of each test dataset. This approach will stop \our's optimization once it reaches optimal performance on this merged validation set, ensuring efficiency and preventing overfitting.

\subsection{Software and hardware details}
We implement \our in Pytorch Geometric framework~\cite{fey_fast_2019}. We conduct our experiments on a Linux system equipped with an NVIDIA A100 GPU with 80GB of memory.

\section{Supplementary experiments}
\subsection{Synthetic graphs}
\begin{table}[h]
    \centering
    \caption{Link prediction results on synthetic Triangular/Grid lattice graphs evaluated by Hits@50. The format is average score ± standard deviation. The top three models are colored by \textbf{\textcolor{red}{First}}, \textbf{\textcolor{blue}{Second}}, \textbf{\textcolor{violet}{Third}}.}\label{tab:syn}
    \resizebox{\linewidth}{!}{%
    \begin{tabular}{lcc}
    \toprule
        & \textbf{Triangular} & \textbf{Grid}\\
    \midrule
    \rowcolor{Gray}
    \multicolumn{3}{c}{Heuristics} \\
    \textbf{CN} & $73.58 {\scriptstyle \pm 0.81}$ & $0.00 {\scriptstyle \pm 0.00}$ \\
    \textbf{AA} & $73.58 {\scriptstyle \pm 0.81}$ & $0.00 {\scriptstyle \pm 0.00}$ \\
    \textbf{RA} & $73.58 {\scriptstyle \pm 0.81}$ & $0.00 {\scriptstyle \pm 0.00}$ \\
    \textbf{PA} & $0.00 {\scriptstyle \pm 0.00}$ & $0.00 {\scriptstyle \pm 0.00}$ \\
    \textbf{SP} & $97.91 {\scriptstyle \pm 0.63}$ & \firstt{86.04}{1.11} \\
    \textbf{Katz} & $90.08 {\scriptstyle \pm 0.67}$ & $56.79 {\scriptstyle \pm 0.99}$ \\
    
    \midrule
    \rowcolor{Gray}
    \multicolumn{3}{c}{SEAL} \\
    \textbf{Supervised} & \firstt{99.29}{0.28} & \secondt{79.45}{1.09} \\
    \textbf{Pretrained Only} & $98.11 {\scriptstyle \pm 0.84}$ & $61.48 {\scriptstyle \pm 0.57}$ \\
    \textbf{Pretrain \& Finetune} & \thirdt{98.35}{0.57} & \thirdt{78.24}{0.79} \\
    \midrule
    \rowcolor{Gray}
    \multicolumn{3}{c}{Ours} \\
    \textbf{UniLP} & \secondt{98.73}{0.49} & $77.39 {\scriptstyle \pm 1.38}$ \\
    \bottomrule
    \end{tabular}
}
\end{table}
We deployed our pretrained \our on the synthetic graph shown in Figure~\ref{fig:syn}, with outcomes presented in Table~\ref{tab:syn}. These findings demonstrate that \our matches the performance of both models that are fully trained on the entire graph and those that undergo explicit finetuning. This performance underscores the efficacy of \our's ICL capability, affirming its ability to dynamically adapt to synthetic graph environments and learn connectivity patterns directly from in-context links without the need for additional training or finetuning.

\subsection{Diversifying context}
\label{app:diverse}
\begin{figure}[h]
\begin{center}
\centerline{\includegraphics[width=\linewidth]{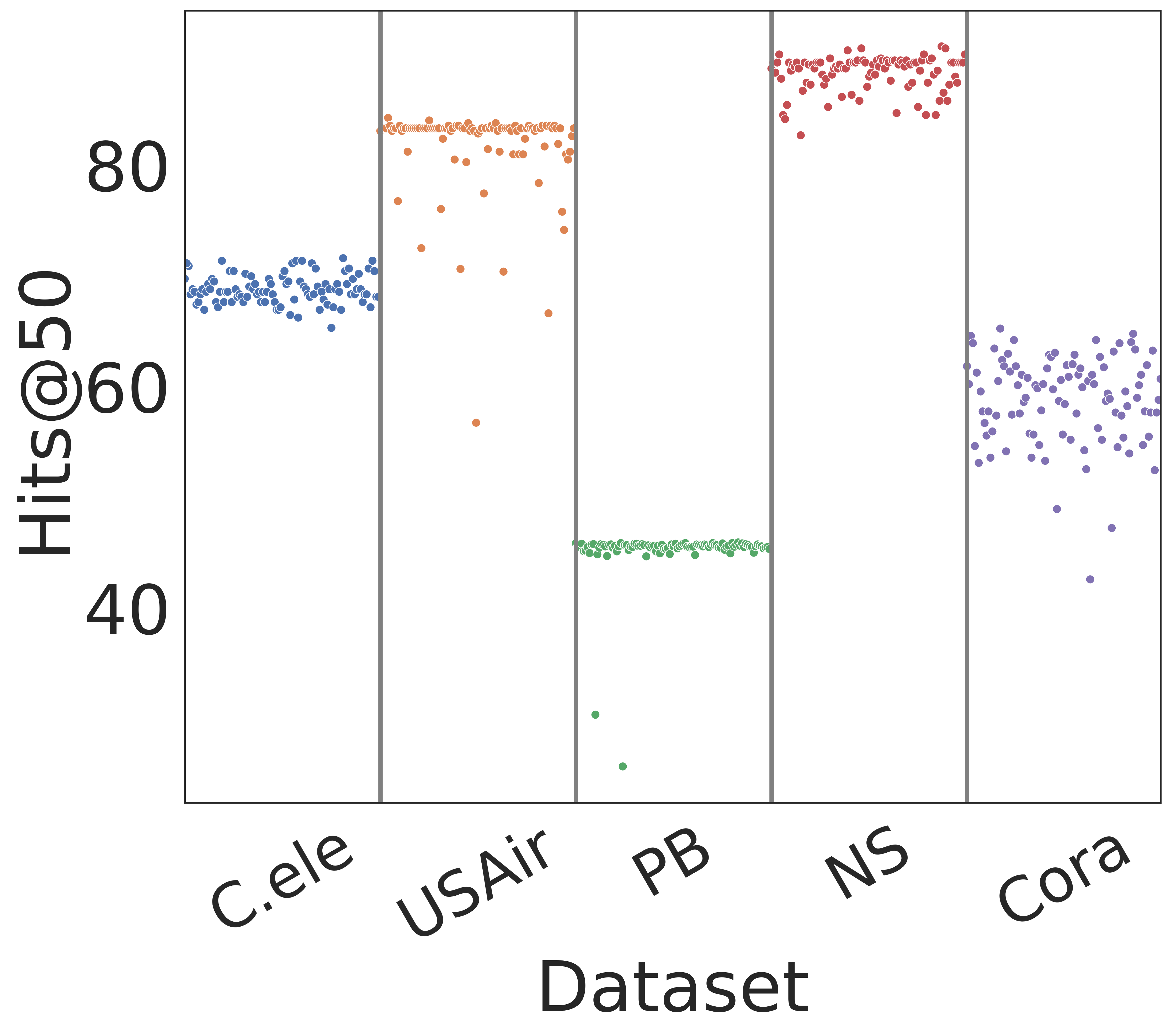}}
\caption{Diverse sets of in-context links on LP performance.}\label{fig:diverse}
\end{center}
\end{figure}

In our prior analysis, we utilized a set of in-context links sampled from each target graph to serve as context. This section delves into the impact of varying these in-context links by employing different random seeds, aiming to discern the sensitivity of \our's performance to the specific composition of in-context links for each target graph. The outcomes of this investigation are detailed in Figure~\ref{fig:diverse}. The findings reveal that the choice of in-context links indeed affects \our's performance across the test datasets to varying extents, highlighting the importance of the selection process for these contextual links in optimizing the model's efficacy. We leave the study on the selection of the in-context links as future work.

\subsection{Varying positive-to-negative ratios of in-context links}
\label{app:ratio}

\begin{figure}[h]
\begin{center}
\centerline{\includegraphics[width=\linewidth]{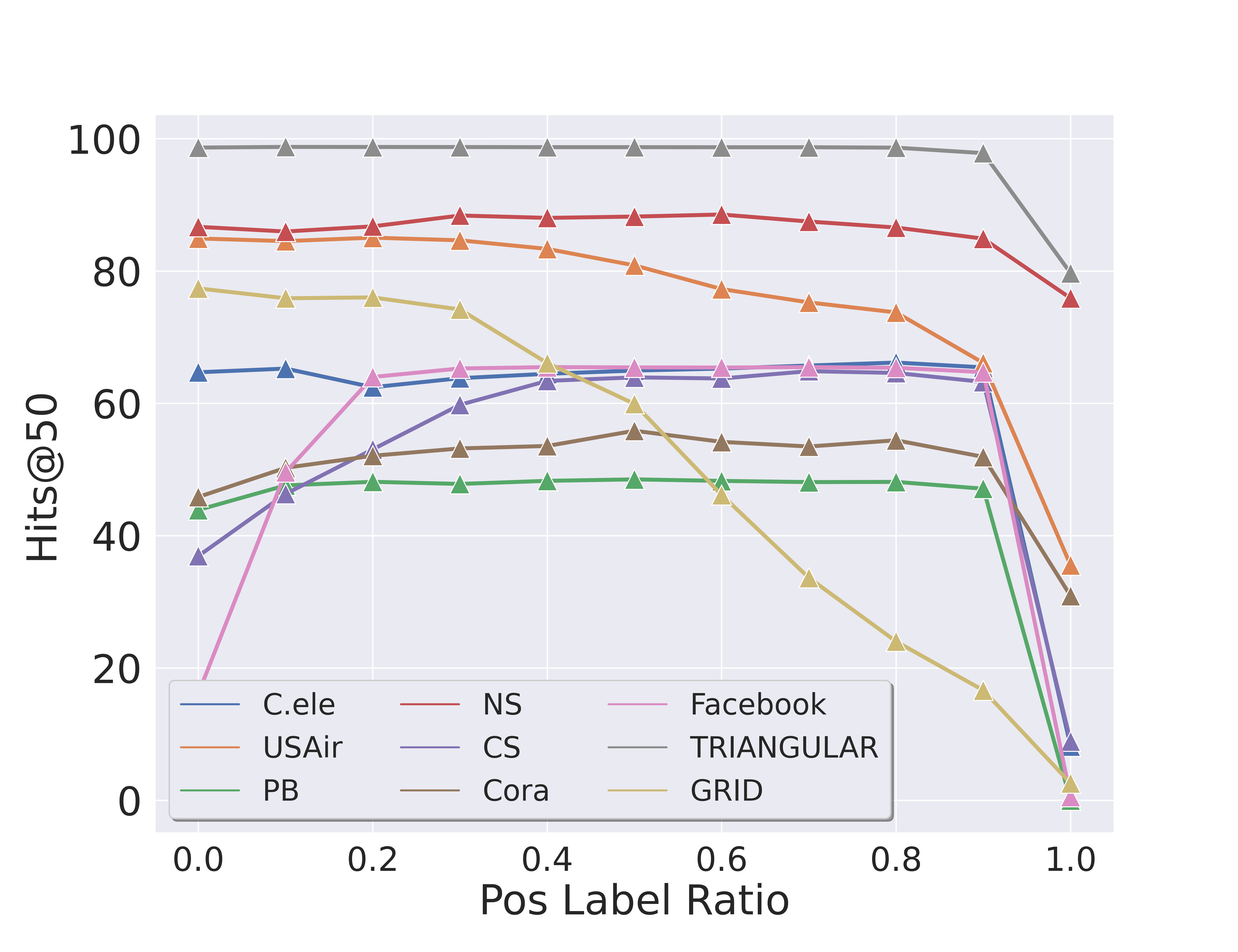}}
\caption{Influence of positive-to-negative in-context link ratios on LP performance.}\label{fig:ratio}
\end{center}
\end{figure}

In our previous exploration of LP performance, we initially maintain a balanced set of positive and negative in-context links for each query link during both pretraining and testing phases. This study delves into the effects of changeing the ratio of positive to negative in-context links, while keeping the total count constant at $200$, to assess the impact on LP accuracy. The findings, shown in Figure~\ref{fig:ratio}, reveal several noteworthy observations.

Remarkably, for the majority of graph datasets examined, increasing the proportion of negative samples—contrary to intuitive expectations—does not detract from performance and, in some cases, matches the efficacy of a balanced distribution of in-context links. This phenomenon indicates that negative samples are equally informative as positive ones for leveraging the ICL capabilities of \our. Specifically, in the case of the synthetic Grid graph, a higher ratio of negative samples significantly enhances LP performance, given a fixed total number of in-context links. This improvement may stem from the symmetry of positive link structures within the Grid graph, which exhibit a consistent connectivity pattern. The introduction of a greater variety of negative samples seems to enrich the model's learning context, effectively harnessing \our's ICL potential to capture more diverse patterns.

An exception to this trend is observed with the Facebook graph dataset, where a balance between positive and negative in-context links yields the most favorable outcomes. This suggests that for certain graph types, a balanced approach to in-context link selection optimizes LP performance.


\section{Theoretical analysis}
\subsection{More discussions about the connectivity patterns}
\label{app:pattern}
In the initial definition (Definition~\ref{def:connectivity}), connectivity patterns are characterized as ordered sequences of events that are satisfied by the features of links. This concept underpins the idea that if two graphs exhibit identical connectivity patterns, a LP model trained on one graph could theoretically be applied to the other without retraining. The rationale behind this is rooted in the LP task's core objective: to prioritize true links over false ones through ranking. Hence, a consistent ranking mechanism across different graphs allows for the same heuristic-based link predictor to be effectively utilized for LP tasks across those graphs.

It might be tempting to equate connectivity patterns directly with graph distributions; however, this is a misconception. Graphs can share identical connectivity patterns yet differ significantly in their underlying distributions. An illustrative example is provided by graphs generated through the Stochastic Block Model~\cite{holland_stochastic_1983} with distinct parameters, which may still present identical connectivity patterns provided their intra-block edge probabilities are higher than those between blocks.

Consequently, despite real-world graphs often exhibiting varied underlying distributions—reflected in aspects such as node degrees, graph sizes, and densities—the question of whether a singular, common connectivity pattern exists across diverse graphs remains non-trivial. This inquiry forms the theoretical foundation for our Universal Link Predictor model, challenging us to explore the feasibility of applying one singular link prediction methodology in a world of inherently distinct graph structures.

\subsection{Proof for Theorem~\ref{thm:conflict}}
We first restate the theorem and proceed with the proof:

Define $A_2=|\pi_2(u,v)|\geq1$ and $A_3=|\pi_3(u,v)|\geq1$ as elements of $ \omega $. The connectivity patterns on Grid and Triangular graphs are distinct. Specifically:\\
(i) On Grid: $ \omega=[A_3,A_2] $;
(ii) On Triangular: $ \omega=[A_2,A_3] $.

\begin{proof}
In a Grid graph, the probability of a connection given a 2-hop simple path, $p(y=1|A_2)$, can be expressed as $\frac{p(y=1,A_2)}{p(A_2)}$. The absence of any 2-hop connected node pairs $(u,v) \in E^o$ implies $p(y=1,A_2)=0$, leading to $p(y=1|A_2)=0$.

Considering the symmetric nature of nodes in a synthetic Grid graph, we select an arbitrary node as an anchor. Identifying nodes with a 3-hop simple path to this anchor reveals that:
\begin{equation}
p(y=1|A_3) = \frac{p(y=1,A_3)}{p(A_3)} = \frac{4}{16} = \frac{1}{4}.
\end{equation}
This calculation confirms the connectivity sequence on Grid as $ \omega=[A_3,A_2] $.

Conversely, in a Triangular graph, the probabilities given a 2-hop and a 3-hop simple path are calculated as:
\begin{align*}
p(y=1|A_2) = \frac{p(y=1,A_2)}{p(A_2)} = \frac{6}{18} = \frac{1}{3},\\
p(y=1|A_3) = \frac{p(y=1,A_3)}{p(A_3)} = \frac{6}{36} = \frac{1}{6}.
\end{align*}
Thus, establishing the connectivity sequence for Triangular as $ \omega=[A_2,A_3] $, which is in direct contrast to that of Grid graphs, highlighting the inherent difference in their connectivity patterns.
\end{proof}

\end{document}